\documentclass{article}
\usepackage{spconf,amsmath,epsfig}

\usepackage{amsfonts,amssymb}
\usepackage{overpic}
\usepackage{graphicx}
\usepackage[caption=false]{subfig}
\usepackage[colorlinks=true, allcolors=blue]{hyperref}
\usepackage[ruled,vlined,linesnumbered,noresetcount]{algorithm2e}
\usepackage{amsmath}
\usepackage{multicol}
\usepackage{multirow}
\DeclareMathOperator{\NN}{NN}
\DeclareMathOperator{\ODESolve}{ODESolve}

\title{Data-Driven Invertible Neural Surrogates of Atmospheric Transmission}

\name{James Koch, Brenda Forland, Bruce Bernacki, Timothy Doster, Tegan Emerson
\thanks{Manuscript accepted for presentation and publication at the 2024 IEEE International Geoscience and Remote Sensing Symposium (IGARSS). This work was conducted under the Laboratory Directed Research and Development Program at PNNL, a multi-program national laboratory operated by Battelle for the U.S. Department of Energy under contract DE-AC05-76RL01830. Contact: \href{mailto:james.koch@pnnl.gov}{james.koch@pnnl.gov}}}
\address{Pacific Northwest National Laboratory \\ Richland, WA}

\begin{document}
%
\maketitle
\begin{abstract}
We present a framework for inferring an atmospheric transmission profile from a spectral scene. This framework leverages a lightweight, physics-based simulator that is automatically tuned -- by virtue of autodifferentiation and differentiable programming -- to construct a surrogate atmospheric profile to model the observed data. We demonstrate utility of the methodology by (i) performing atmospheric correction, (ii) recasting spectral data between various modalities (e.g. radiance and reflectance at the surface and at the sensor), and (iii) inferring atmospheric transmission profiles, such as absorbing bands and their relative magnitudes. 
\end{abstract}
\section{Introduction}
\label{sec:intro}

\textit{Data-Driven Invertible Neural Surrogates of Atmospheric Transmission}, or DINSAT, is a modeling framework for inferring atmospheric transmission profiles suitable for atmospheric correction and transmission modeling tasks. This framework is built upon the notion of tunable differential equations (i.e. Neural Ordinary Differential Equations \cite{chen2018neural}) that qualitatively reproduce the dominant physics present in atmospheric transmission and spectral sensing. 

In remote sensing, the process by which sensor measurements are converted to surface reflectance is termed \textit{atmospheric correction}. In the processing pipeline for downstream analysis tasks, atmospheric correction introduces the most uncertainty \cite{hadjimitsis2009use}: the atmosphere interacts with radiation nonlinearly (e.g. adjacency effects or nonlinear transmission profiles) which limits the performance of many in-scene correction routines. Such methods include dark background subtraction \cite{teillet1995dark} or Quick Atmospheric Correction (QUAC) \cite{bernstein2012quick} which rely upon statistics calculated from a particular scene to determine an appropriate data transform. While limited in accuracy, these methods are low-cost and have proven effectiveness. For situations where higher fidelity corrections are required, methods based upon radiative transfer simulations can be used, such as the Fast Line-of-Sight Atmospheric Analysis of Spectral Hypercubes (FLAASH) \cite{anderson2002modtran4}, which leverages the MODerate resolution atmospheric TRANsmission code (MODTRAN) \cite{berk2014modtran}. Such methods perform best when situational properties of a spectral scene are known; e.g. atmospheric properties, illumination setting, surface temperature, distance to the sensor, and the specific sensor architecture. 

Machine learning (ML)-based methods excel at learning relationships between data distributions. An ML model with enough parameters may be able to adequately capture data distributions in different domains (e.g. reflectance and radiance units, at-sensor and atmospherically-corrected measurements, etc.) with conventional architectures such as convolutional neural networks, autoencoders, and transformers. Examples include using Gaussian processes \cite{estevez2020gaussian,basener2023gaussian} and Deep Neural Networks (DNNs) for estimating the transformation between the associated data domains \cite{xu2020multiple,pyo2020integrative,qamar2023atmospheric}. Recent advances in generative modeling have also enabled novel architectures for removing the effects of an atmosphere on a scene, such as in Stelter and Sundberg \cite{stelter2023diffusion}. 

There are drawbacks and risks with conventional ML models in this setting. The complex physical relationships, such as atmospheric transport, can require highly-parameterized models which are accompanied by high training burdens. In addition, such methods may fail to generalize beyond what has been seen during training (out-of-distribution data). Lastly, in many situations it may be advantageous to perform analysis in either units of reflectance or radiance, in which case one may seek to leverage an atmospheric transmission model to add such effects to a scene. This requires \textit{invertibility} of the atmospheric correction model. Many current architectures only operate in a single direction; i.e. they are not invertible. Separate models may need to be trained for each transformation of the data.

Here, we seek to construct minimally-parameterized models for atmospheric correction that are (i) invertible, (ii) physically consistent, (iii) robust to out-of-distribution data, and (iv) have minimal data (at-sensor radiance) and meta-data (lighting, geometry, temperature, etc.) requirements for training. This work builds upon and formalizes ideas presented in Koch et al. \cite{koch2023neural}.

\begin{figure}[]
\centering
        \begin{overpic}[width=0.6\columnwidth]{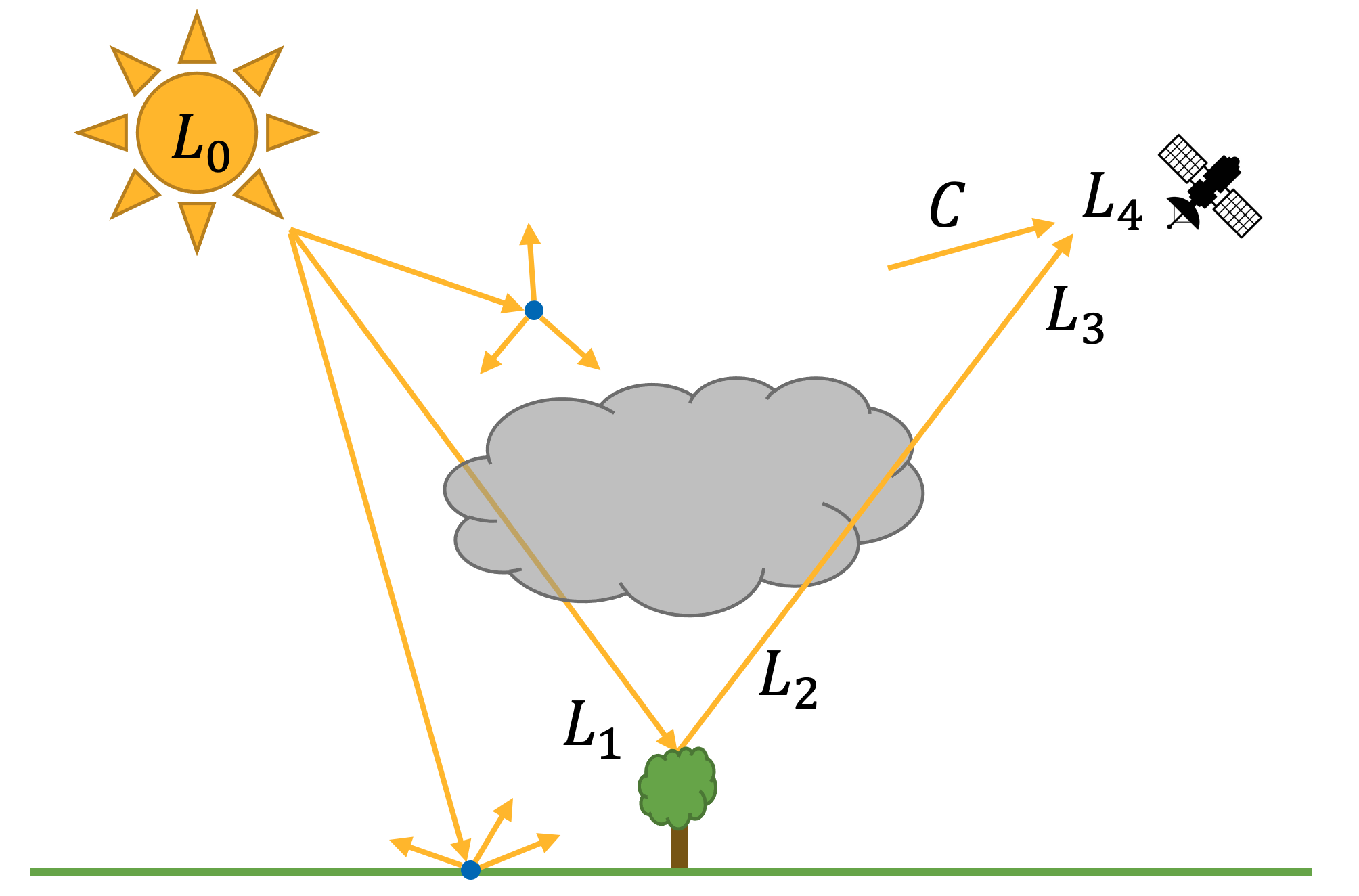}
        \end{overpic}  
        \caption{Solar radiation, denoted $L_0$, is both absorbed and scattered in the atmosphere, the extent of which is tightly coupled with atmospheric properties. A sensor records radiation intensity as a combination of lumped upwelling effects ($C$) and direct path radiation ($L_3$).}
        \label{fig:steps}
\end{figure}

\section{DINSAT Methodology} \label{sec:methods}

A notional spectral scene is depicted in Fig. \ref{fig:steps}. Electromagnetic (EM) radiation is emitted by the sun ($L_0$) and is absorbed and scattered as it travels through the atmosphere. This radiation ($L_1$) reflects off of a surface target and is transmitted back through the atmosphere to a sensor. The total at-sensor radiation is the combination of path radiance ($L_3$) and an offset ($C$) from diffuse surface reflection, multiple scattering, etc. The at-sensor measurement is the composition of these processes: $L_4 = C + T\left(\rho T\left(L_0\right) \right)$, where $T(\cdot)$ is the \textit{atmospheric transmission operator} and $\rho$ is surface reflectance. The physics of the interaction of EM radiation with a transmissive medium is governed by Maxwell's equations -- the solution of which is intractable for our task. We leverage the knowledge that radiation is absorbed and transmitted consistent with the Beer-Lambert Law (exponential decay). We construct the transmission operator as the solution to the tunable Ordinary Differential Equation (ODE) $\frac{dL}{dx} = f\left(L ; \theta\right)$:
\begin{equation} \label{eq:ode}
    T\left( L \right) := L + \int_{x_0}^{x_{end}} f\left( L;\theta \right) dx,
\end{equation}
where $x$ is the integration distance; i.e. the path length of radiation through a transmissive medium, $f$ are the dynamics imparted on the radiation by the medium, which depends parametrically on $\theta$. The restriction on the form of this ODE is that $f:\mathbb{R}^n \rightarrow \mathbb{R}^n_{\leq 0}$ such that dissipation through transmission is enforced by construction. The integration distance is set to 1 for all analyses in this work.

\subsection{Neural Ordinary Differential Equations}
Neural ODEs \cite{chen2018neural} evolve a system state $L$ through an independent variable $x$ according to a learnable or otherwise tunable right-hand-side. In the case of a black-box Neural ODE, this is typically a standard multi-layer perceptron (MLP): $f(L;\theta) := \NN (L;\theta)$, where $\NN\left(\cdot\right)$ is a MLP. Evolving $L$ is the equivalent of solving the associated IVP. The numerical solution of the IVP can be performed with any number of standard ODE solvers, such as forward Euler or another higher-order Runge-Kutta scheme. We denote the numerical solution to the IVP in Eq. \ref{eq:ode} as:
\begin{equation} \label{eq:odesolve}
    L\left(x_{end}\right) = \ODESolve \left( f\left( L;\theta \right), L\left(x_0\right), \left[x_0, x_{end} \right]\right),
\end{equation}
where $\ODESolve\left(\cdot\right)$ is a numerical integration scheme. The $\ODESolve(\cdot)$ can be made differentiable through backpropagating through the elementary operations of the solver (e.g. differentiable programming) or via an adjoint state method \cite{chen2018neural}. The mapping $f$ is tuned by obtaining gradients of a loss function with respect to the parameters contained in $\theta$ and passing this gradient information to an optimization routine. Specific functional forms or domain-informed terms can be imparted on a model by modifying Eq. \ref{eq:ode} to contain this information by construction (the Universal Differential Equations paradigm \cite{rackauckas2020universal}).

\subsection{Atmospheric Correction}
Our goal is to obtain pixel-wise estimates for surface reflectance from top-of-atmosphere hyperspectral images. To do so, the following are needed: data in a consistent unit set, an estimate of background radiation (an assumed constant offset $C$), and an algebraic representation of surface reflectance to be used in a loss function and optimization problem. 

The constant $C\in \mathbb{R}^n_{\geq 0}$ is set to the minimum value of each band (dark background subtraction). If available, the known incident radiation $L_0 \in \mathbb{R}^n_{\geq 0}$ can be used to rescale radiance data into the range $[0,1]$. However, in many situations, the incident radiation is either unknown or poorly quantified (e.g. estimates through blackbody curves and estimates of the geometry of the spectral scene). For these situations, we define the incident solar spectrum to be a flat spectrum of constant magnitude; $m\in \mathbb{R}_{>0}$: $L_0 = m\times \mathbf{1}_n = m\times(1,1,...,1)$ such that $L_4/m \in [0,1]$.

Surface reflectance can be derived as $\rho = \frac{L_2}{L_1}$, as shown in Fig. \ref{fig:steps}. These terms can be estimated through the use of the transmission operator and its inverse:
\begin{equation} \label{eq:atm_correction}
    \rho = \frac{L_2}{L_1} = \frac{ T^{-1}\left(\frac{L_4 - C}{m} \right)}{T\left( \mathbf{1}_n \right)}.
\end{equation}

Evaluating this calculated reflectance against known targets (Section \ref{sec:sup}) or other metrics (Section \ref{sec:unsup}) allows one to obtain error and gradient information needed to automatically tune the transmission operator in an optimization framework.

\section{Demonstration} \label{sec:experiments}

The \textit{Target Detection Blind Test Project} \cite{snyder2008development} is a dataset and online platform for testing and evaluating hyperspectral target detection algorithms. The dataset consists of overhead imagery of Cooke City, Montana captured by the HyMap sensor. The spectral scene contains 126 spectral bands in the 450 - 2500nm range with $3\times3$ meter pixel spatial resolution. Seven targets of known spectral signatures were placed within the scene for calibration panel purposes. Here, we leverage this dataset to evaluate atmospheric correction algorithms in the ability to reconstruct surface reflectance.

\begin{figure}[]
\centering
        \begin{overpic}[width=1\columnwidth]{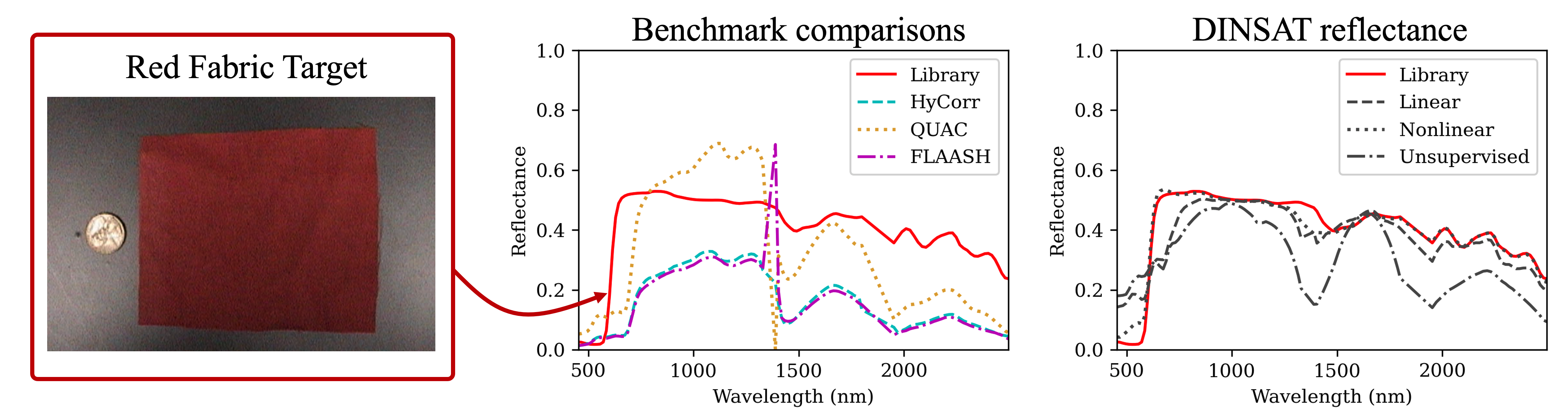}
        \put(2,26){(a)}
        \put(35,26){(b)}
        \put(70.5,26){(c)}
        \end{overpic}  
        \caption{An example in-scene target is shown in (a) (reproduced from Snyder et al. \cite{snyder2008development}). The laboratory-measured spectral signature for the target is shown in (b) alongside estimates provided by off-the-shelf atmospheric correction routines. In (c), shown are variations of our model-derived corrected spectra compared against the same reference spectra.}
        \label{fig:baseline_refl}
\end{figure}

The targets present in the scene are four fabrics of different colors and materials and three vehicles of different colors, makes, and models. The training dataset provides Regions of Interest (ROIs) associated with each of the targets. As benchmarks, we perform atmospheric correction on the scene using standard off-the-shelf tools: (i) HyCorr (as packaged with the dataset), (ii) QUAC, and (iii) FLAASH. Figure \ref{fig:baseline_refl}(a) shows an example fabric patch used as a target in the scene. The corresponding laboratory-measured reflectance is shown in Fig. \ref{fig:baseline_refl}b alongside the benchmark comparisons (averaged over the ROI corresponding to the target).

\begin{figure}[!ht]
    \centering

    \subfloat[Supervised: Linear transmission profile (ensemble of models)]{
        \begin{overpic}[width=1.0\columnwidth]{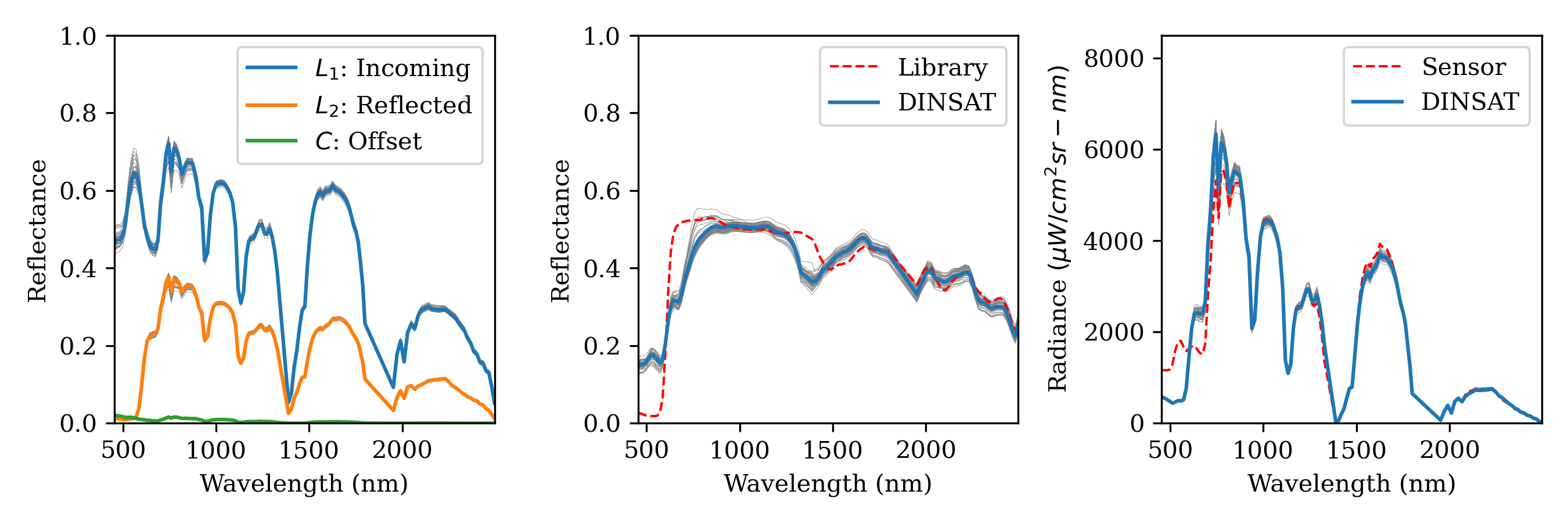}

        \end{overpic}     
    }
    \vspace{-0.1in}
    \subfloat[Supervised: Nonlinear transmission profile]{
        \begin{overpic}[width=1.0\columnwidth]{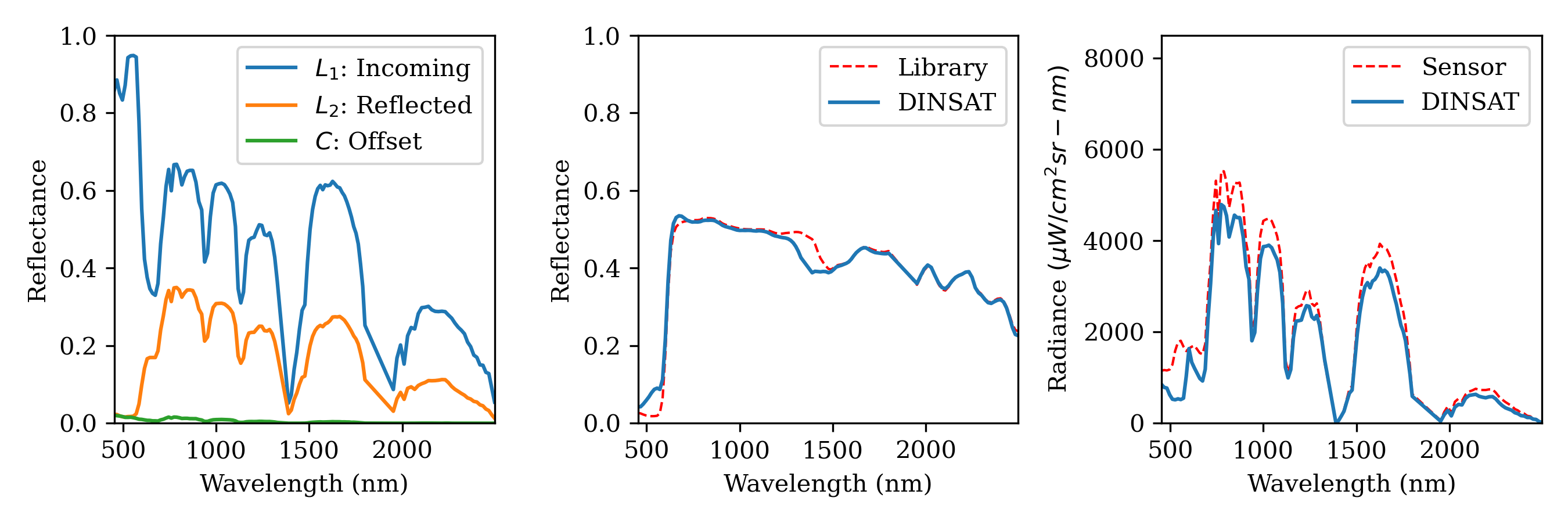}
        \end{overpic}     
    }
    \vspace{-0.1in}
    \subfloat[Unsupervised: Linear transmission profile (ensemble of models)]{
        \begin{overpic}[width=1.0\columnwidth]{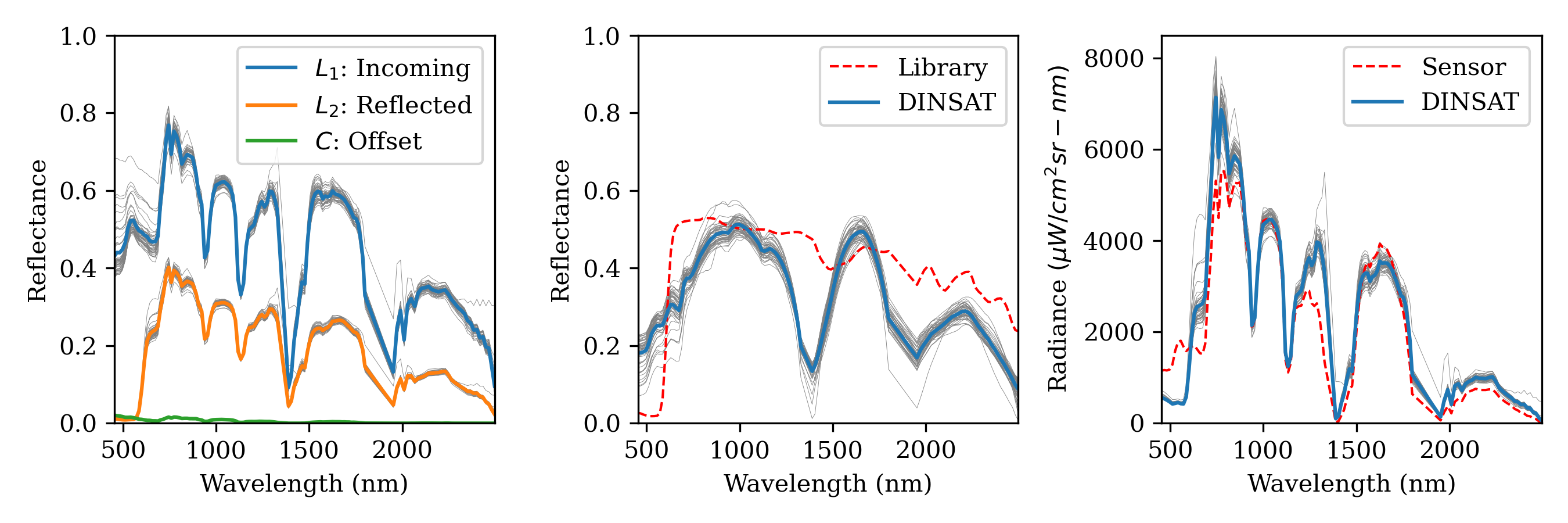}
        \end{overpic}     
    }
    \vspace{-0.1in}
    \caption{Summary of DINSAT results. Left: estimated in- and outgoing radiation at a surface-based target. Middle: Estimated reflectance spectra. Right: estimates of at-sensor radiance given a DINSAT model and known (library) spectra.}
    \label{fig:results}
    
\end{figure}

\subsection{Supervised Models}\label{sec:sup}
We demonstrate supervised training for: (i) a linear transmission profile, and (ii) a nonlinear transmission profile. For these tasks, training and testing datasets comprised of per-pixel ($p$) pairwise at-sensor radiance, denoted $L_{4,p}$, and ground-truth surface reflectance, denoted $\rho_p$. The ground-truth reflectance spectra paired with each $L_{4,p}$ is the reference spectra associated with the ROI as packaged by the Target Detection Blind Test Program. For this study, a 24, 6, 70\% split between train, validation, and test data is employed, corresponding to 35, 9, and 101 pixels contained in each data split, respectively. We construct two atmospheric transmission profiles to test:

\begin{equation} \label{eq:linear}
    \frac{dL}{dx} = -\alpha L, \text{and}
\end{equation}
\begin{equation} \label{eq:nonlinear}
    \frac{dL}{dx} = h(g(L;\theta);\phi) L .
\end{equation}
Equation \ref{eq:linear} simply defines the evolution of the spectra to decay exponentially as a function of transmission distance (with $\alpha \in \mathbb{R}^n_{\geq 0}$). In Eq. \ref{eq:nonlinear}, nonlinear effects are included by defining an autoencoder-decoder structure, with $g: \mathbb{R}^n \rightarrow \mathbb{R}^q$ and $h: \mathbb{R}^q \rightarrow \mathbb{R}^n$. The mappings $g$ and $h$ are constructed as MLPs, each with one hidden layer of size 12 and sigmoid-activated. The latent dimensionality $q$ is set to 3. 

The loss function for the optimization problem is a combination of Mean Squared Error (MSE) between predicted and true reflectance and a Finite Difference (FD) regularization; $\mathcal{L} = \mathcal{L}_\text{MSE} + \lambda \mathcal{L}_\text{FD},$ with:
\begin{equation} \label{eq:sup_loss}
    \begin{split}
        \mathcal{L}_\text{MSE} &= \frac{1}{N_p N_b}\sum_{p=1}^{N_p} \sum_{i=1}^{N_b} \left( \rho^{(p)}_i - \hat{\rho}^{(p)}_i \right)^2, \\
        \mathcal{L}_\text{FD} &= \\ 
        \frac{1}{N_p (N_b-1)}\sum_{p=1}^{N_p} &\sum_{i=1}^{N_b-1} \left( \left(\rho^{(p)}_{i+1} - \rho^{(p)}_{i} \right) - \left(\hat{\rho}^{(p)}_{i+1}- \hat{\rho}^{(p)}_{i} \right) \right)^2, 
    \end{split}
\end{equation}
where $N_p$ is the number of pixels in the dataset, $N_b$ is the number of bands in the spectrum, $\lambda$ is a scalar hyperparameter specifying the relative weight of the two loss terms, and the notation $\hat{\cdot}$ denotes model output. For both cases, $\lambda = 1$ and model tuning is performed with the Adam optimizer with a learning rate of 0.01 until a convergence criteria is met. 

The linear transmission profile is inexpensive to train and deploy on this particular scene of size  $800\times280\times126$ (on the order of 1-3 minutes per model instance for training and evaluation on a single CPU thread of a mid-grade laptop). For these cases, an ensemble of 50 numerical experiments were constructed and run with different data shuffles of the training data. For the same conditions, the nonlinear transmission model is about two orders of magnitude more computationally expensive. Only one trained nonlinear model is presented herein.

Figure \ref{fig:results}(a) profiles the ensemble of learned linear transmission models. In the left column of Fig. \ref{fig:results}, the estimated incident and reflected spectra are shown with averages overlaid. Model-derived surface reflectance is shown in the middle column. The right column corresponds to sending the library spectrum for the red fabric through the forward pass of the model to obtain an estimate for the at-sensor radiance measurement. Figure \ref{fig:results}(b) details the learned profiles for the nonlinear transmission case in the same manner. For comparison, the percent Mean Squared Error (MSE) corresponding to these spectra for surface reflectance is 1.5\% and 0.3\% for the linear and nonlinear cases, respectively. Similarly, the errors for the at-sensor radiance estimates are 0.6\% and 16\%. 

\subsection{Unsupervised Model} \label{sec:unsup}
Here, only at-sensor radiance data is available and additional regularization is required to promote convergence. We exclusively examine the minimally-parameterized linear transmission profile (Eq. \ref{eq:linear}). The loss function is $\mathcal{L} = \lambda_1 \mathcal{L}_\rho + \lambda_2 \mathcal{L}_\text{T} + \lambda_3 \mathcal{L}_\text{FD}$ with:
\begin{equation} \label{eq:unsup_loss}
    \begin{split}
        \mathcal{L}_\rho &= \frac{1}{N_p N_b}\sum_{p=1}^{N_p} \sum_{i=1}^{N_b} \hat{\rho}^{(p)}_i, \\
        \mathcal{L}_\text{T} &= \frac{1}{N_b}\sum_{i=1}^{N_b} T(\mathbf{1})_i, \\
        \mathcal{L}_\text{FD} &= \frac{1}{N_p (N_b-1)}\sum_{p=1}^{N_p} \sum_{i=1}^{N_b-1} \lvert\hat{\rho}^{(p)}_{i+1}- \hat{\rho}^{(p)}_{i}\rvert. \\
    \end{split}
\end{equation}
$\mathcal{L}_\rho$ penalizes the mean reflectance spectrum, $\mathcal{L}_\text{T}$ penalizes the transmission of the medium, and $\mathcal{L}_\text{FD}$ penalizes the slope of the estimated reflectance spectrum. $\mathcal{L}_\rho$ and $\mathcal{L}_\text{T}$ both seek to maximally dissipate energy. For this experiment, $\lambda_1, \lambda_2$, and $\lambda_3$ are $10^{-2}$, $10^{-2}$, and 1.0, respectively.

A diverse set of pixels from the spectral scene are needed to construct a well-behaved loss landscape; in our example here, we randomly selected 0.05\% of the pixels (112) to constitute our dataset, 87 for the train split and 25 for test split. Training was performed in the same manner as the experiments presented in Section \ref{sec:sup} with an ensemble of 50 models, each with randomly-sampled training and test pixels. The trained model profiles are given in Fig \ref{fig:results}(c). The salient spectral features in the transmission spectra and in the reflectance spectra remain, however the error between ground-truth reflectance and DINSAT-predicted surface reflectance (averaged spatially over the ROI and over the ensemble of runs) increased to 16\% MSE. Similarly, the MSE for the estimated at-sensor radiance measurement grew to 11\%.

\section{Conclusion} \label{sec:discussion}
Our presented methodology -- DINSAT -- is a novel framework for performing atmospheric correction by learning surrogates of atmospheric transmission profiles from data. We have shown this method to be comparable to many commercial-off-the-shelf methods while requiring less supervision, and in some cases, no supervision. Here, we use DINSAT to provide an estimate for surface reflectance, but in general, we can reformulate the problem to achieve a different task (e.g. reformulating the forward pass to predict at-sensor radiance directly) or to couple with a downstream task (e.g. a deep learning-based detector or classifier). 

This method can be extended to include other sensing paradigms. For example, measurements of underwater reefs through a satellite-based sensing platform would need both atmospheric correction as well as water column correction. Being both composable and differentiable, this framework can be readily extended to model multi-media (e.g. atmosphere and water) or spatially-varying transmission profiles. Additionally, the flexibility of the software implementation allows for the inclusion of more complex physics. Examples include (i) learning adjacency effects, both spatially and spectrally, (ii) the inclusion of stochastic processes (i.e. Stochastic Differential Equations), and (iii) time varying atmospheric profiles.

\bibliographystyle{IEEEbib}
\bibliography{report}

\end{document}